\documentclass{article}

     \PassOptionsToPackage{numbers, compress}{natbib}
     \usepackage[final]{neurips_2019}

\usepackage[utf8]{inputenc} 
\usepackage[T1]{fontenc}    
\usepackage{hyperref}       
\usepackage{url}            
\usepackage{booktabs}       
\usepackage{amsfonts}       
\usepackage{nicefrac}       
\usepackage{microtype}      

\usepackage{graphicx} 
\usepackage{amsmath}

\title{Disentangling Interpretable Generative Parameters of Random and Real-World Graphs}

\author{
  Niklas Stoehr, Emine Yilmaz\\
  University College London\\
  \texttt{n.stoehr@outlook.com} \\
  \texttt{emine.yilmaz@ucl.ac.uk} \\
  \And
  Marc Brockschmidt, Jan Stuehmer\\
  Microsoft Research Cambridge\\
  \texttt{mabrocks@microsoft.com} \\
  \texttt{jastueh@microsoft.com} \\
}

\begin{document}

\maketitle

\begin{abstract}
While a wide range of interpretable generative procedures for graphs exist, matching observed graph topologies with such procedures and choices for its parameters remains an open problem. Devising generative models that closely reproduce real-world graphs requires domain knowledge and time-consuming simulation. While existing deep learning approaches rely on less manual modelling, they offer little interpretability. This work approaches graph generation (decoding) as the inverse of graph compression (encoding). We show that in a disentanglement-focused deep autoencoding framework, specifically \textit{$\beta$-Variational Autoencoders ($\beta$-VAE)}, choices of generative procedures and their parameters arise naturally in the latent space. Our model is capable of learning disentangled, interpretable latent variables that represent the generative parameters of procedurally generated random graphs and real-world graphs. The degree of disentanglement is quantitatively measured using the \textit{Mutual Information Gap (MIG)}. When training our $\beta$-VAE model on \textit{ER random graphs}, its latent variables have a near one-to-one mapping to the ER random graph parameters $n$ and $p$. We deploy the model to analyse the correlation between graph topology and node attributes measuring their mutual dependence without handpicking topological properties. To allow experimenting with the code, we provide an \href{https://colab.research.google.com/drive/1M--YX4dOSt3imDPdecPbjVX-T6Ae0_OG}{interactive notebook}\footnote{\url{https://colab.research.google.com/drive/1M--YX4dOSt3imDPdecPbjVX-T6Ae0_OG}}.
\end{abstract}

\section{Introduction}
\paragraph{Motivation and Related Work}
Conventional network analysis aims at finding interpretable models that explain interaction dynamics by examining graphs as discrete objects \cite{barabasi2016network}. Random graph generator models \cite{chakrabarti2006graph} like \textit{Erdős–Rényi random graphs (ER graphs)} \cite{erdos1959on} are usually too generic to accurately represent the versatile linking patterns of real-world graphs \cite{chakrabarti2006graph, leskovec2010kronecker, dorogovtsev2002evolution}. Devising models that reproduce characteristic topologies prevalent in social \cite{leskovec2007graph}, biological \cite{vazquez2003a}, internet \cite{zhou2004accurately} or document \cite{menczer2004evolution} graphs typically requires a thorough understanding of the domain and time-consuming graph simulations, thereby imposing strong assumptions and modelling bias. Recently, deep learning on non-euclidean data such as graphs has received substantial attention \cite{bronstein2017geometric}.
As these techniques require little or no explicit modelling and capture complex graph structure \cite{brugere2018network, wang2016structural}, we propose to use them as a tool to obtain interpretable generative parameters of graphs. As a limiting factor, most existing models generate graphs sequentially based on concatenations of node embeddings. These are not only non-interpretable but also impose an artificial node ordering instead of considering a global representation of the entire graph \cite{johnson2017learning, you2018graphrnn, li2018learning, liu2018constrained, li2016gated, simonovsky2018graphvae}. \textit{DisenGCN} \cite{ma2019disentangled} focuses on interpretability, but is limited to node-level linking mechanisms. The latent space of \textit{NetGAN} \cite{bojchevski2018netgan} reveals topological properties instead of generative parameters. Some recent works on interpretable graph embeddings \cite{wang2016structural, grover2016node2vec, cao2016deep, walklets2016perozzi, noutahi2019towards} provide visualizations for inspection, but no parameters suitable for a generative model.

In other domains, interest in model interpretability has caused a focus on the latent space of neural models~\cite{navlakha2017learning}.
Intuitively, the aim is to shape the latent space such that the euclidean distance between the latent representations of two data points corresponds to a ``distance'' between the actual data points~\cite{bengio2013representation}. Latent variables describe probability distributions over the latent space. The goal of latent variable disentanglement can be understood as wanting to use each latent variable to encode one and only one data property in a one-to-one mapping \cite{chen2018isolating}, making the latent space more interpretable. Varying one latent variable should then correspond to a change in one observable factor of variation in the data, while other factors remain invariant~\cite{bengio2013representation}. Most work in this field has been focused on visual and sequential data \cite{chen2018isolating, chen2016infogan, kim2018disentangling, stuehmer2019independent}.

\begin{figure}[t]
  \centering
  \includegraphics[width=1.0\textwidth]{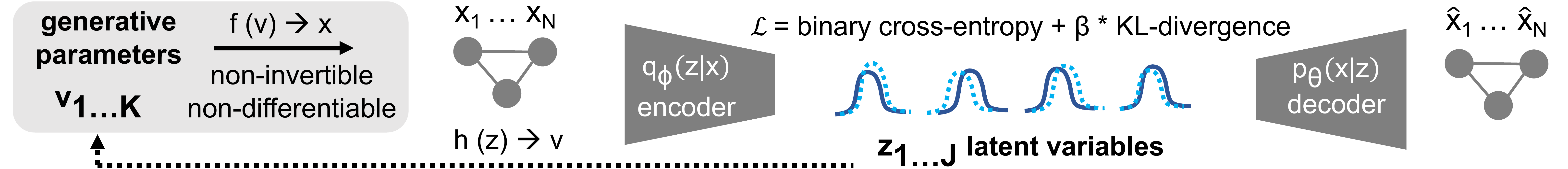}
  \caption{\textbf{Architecture Overview}: We seek a continuous function $h$ mapping the disentangled latent variables $z_j$ into mutually independent, interpretable generative parameters $v_k$.}
  \label{fig:setting}
\end{figure}

\paragraph{Contributions}
We assume that graphs are generated by superposition of interpretable, generative procedures parameterized by generative parameters $v_k$ such as $n$ and $p$ in ER graphs. We hypothesize that these generative parameters $v_k$ can be encoded by a minimal set of disentangled latent variables $z_j$ in an unsupervised machine learning model. To this end, we apply the idea of \textit{$\beta$-Variational Autoencoders ($\beta$-VAE)}~\cite{higgins2017beta} in the context of graphs. Intuitively, our autoencoder tries to compress (encode) a graph into a latent variable representation suitable for generating (decoding) it back into the original graph as outlined in figure \ref{fig:setting}. If the number of latent variables is lower than the dimensionality of the input data, they force a compressed representation that prioritizes the most salient data properties. In this article, we
\\
(1) discuss how to adapt the $\beta$-VAE model to graphs in section \ref{sect:model},\\
(2) apply it to recover parameters for topology-generating procedures in section \ref{sect:eval-topology}, and \\
(3) leverage it to quantify dependencies between graph topology and node attributes in section \ref{sect:eval-attributes}.

\section{Model}
\label{sect:model}
We instantiate the idea of $\beta$-VAEs~\cite{higgins2017beta} with graph-specific encoders and decoders. Our encoder model $q_{\Phi}(z \mid x)$ is a \textit{Graph Convolutional Network (GCN)}~\cite{kipf2017semi} and the decoder $p_{\Theta}(x \mid z)$ is a deconvolutional neural network. Hence, in our setting the encoder is operating on the graph structure, whereas the decoder produces a graph by computing an adjacency matrix. We train this autoencoder in the $\beta$-VAE setting, in which the loss to minimize is $\mathbb{E}_{z \sim q_{\Phi}(z \mid x)}[ \log p_{\Theta}(x \mid z) ] + \beta(\mathbb{KL}(q_{\Phi}(z \mid x) \;\|\; p(z))).$ In the loss term, the reconstruction loss is balanced with the KL regularization term using a parameter $\beta \geq 1$. A higher value of $\beta$ yields stricter alignment to the Gaussian prior $p(z) = \mathcal{N}(0, 1)$, leading to an orthogonalization of the encoding in $z$ \cite{chen2018isolating, chen2016infogan, kim2018disentangling}. To further enforce disentangled representations of $v_k$, we attach an additional \textit{parameter decoder} $h$ to the latent space that learns a direct mapping $h(z) \rightarrow v$ between latent variables $z_j$ and generative parameters $v_k$. If $h$ is implemented as a linear mapping, the latent space needs to align with the generative parameters $v_k$, hence further favoring the result of the encoder $q_{\Phi}(z \mid x)$ to be disentangled. If the latent space is perfectly disentangled, there should exist a one-to-one, bijective mapping $h(z) \rightarrow v$ between latent variables $z_j$ and generative parameters $v_k$. 

For graphs of which we know the ground truth generative parameters $v_k$, we use the metric \textit{Mutual Information Gap (MIG)} \cite{chen2018isolating} to quantify the degree of correlation between $z_j$ and $v_k$. MIG measures both, the extent to which latent variables $z_j$ share mutual information with generative parameters $v_k$, and the mutual independence of the latent variables from each other. The metric ranges between 0 and 1, where 1 represents a perfectly disentangled scenario in which there exists a deterministic, invertible one-to-one mapping between $z_j$ and $v_k$. MIG is computed by first identifying the two latent variables $z_j$ of highest \textit{mutual information (MI)} with each generative parameter $v_k$. The MIG score is then defined as the difference (gap) between the highest and second highest MI, averaged over the generative factors $v_k$.

\section{Evaluation}
\subsection{Modelling Graph Topology with Latent Variables}
\label{sect:eval-topology}

\begin{figure}[h]
\centering
\includegraphics[width=1.0\textwidth]{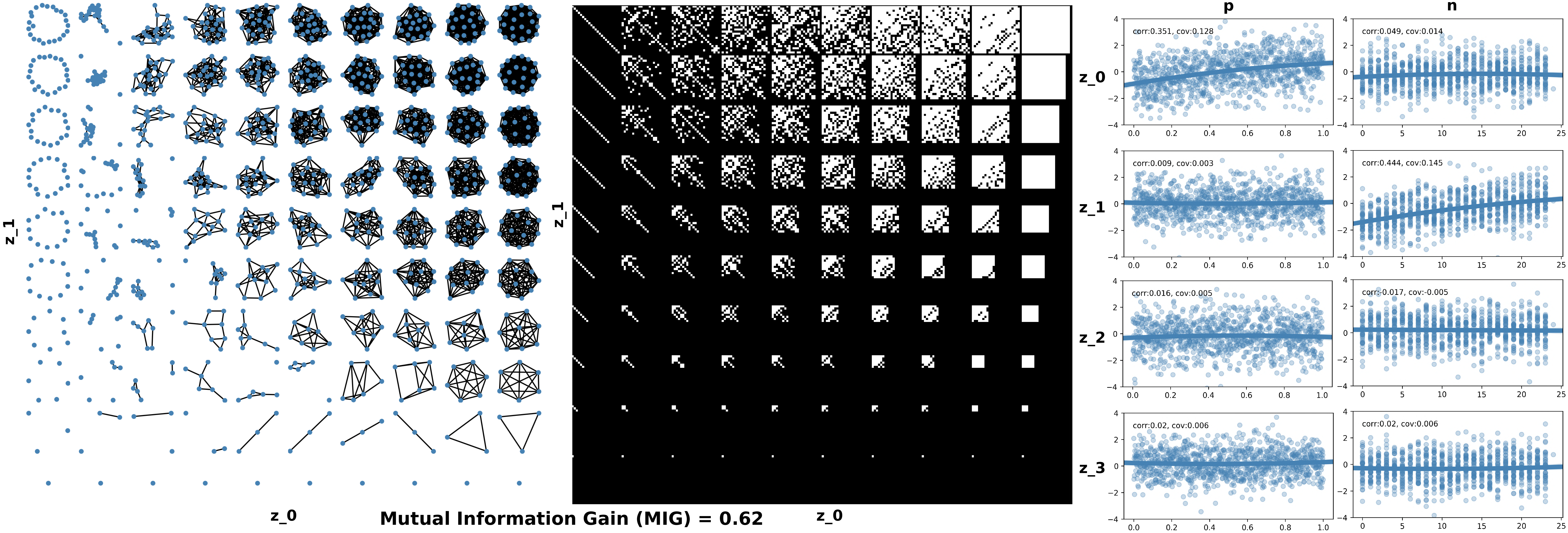}
\caption[Disentangled latent representation of ER graphs]{\textbf{Disentangled latent representation of ER graphs} \quad The latent space appears axis-aligned with $z_0$ and $z_1$ orthogonally representing $p$ and $n$. Changing one latent variable $z_0$ or $z_1$ corresponds to a change in one generative parameter $p$ or $n$ respectively, while being relatively invariant to changes in other parameters. $z_2$ and $z_3$ are not utilized by the model.}
\label{fig:figure_1}
\end{figure}

First, we evaluate our approach on synthetically generated graphs, concretely, ER graphs~\cite{erdos1959on}. The ER generation procedure takes two parameters: the number of nodes $n$ and a uniform linking probability $p$. Ideally, our model should be able to single out these independent generative parameters by utilizing only two latent variables that describe a one-to-one mapping. To test this hypothesis, we generate 10,000 ER graphs, $n$ varying between 1 and 24 and $p$ between 0 and 1. We use these to train our model with a latent space of size $J = 4$ and $\beta = 5.0$. 

To inspect the latent space of the trained model, we sample from $z_j$ in fixed-size steps and decode the sample through $p_{\Theta}(x \mid z)$. Figure \ref{fig:figure_1} shows graphs (on the left) and adjacency matrices (in the center) sampled from the latents $z_0$ and $z_1$ while keeping other latents $z_3, z_4$ fixed. The adjacency matrices allow reading off topological properties of the graphs such as degree distribution and assortativity since nodes are sorted according to the extended \textit{BOSAM}~\cite{guo2006bosam} algorithm. Instead of decoding samples, we may also encode graph instances $x$ with known ground truth generative parameters $v_k$ and observe the latents $z_j$. We generate a new set of 1,000 ER graphs with varying $v_k$ and feed these graphs to the trained model. In figure \ref{fig:figure_1} (on the right), each row displays samples from one latent variable $z_j$ and the columns represent generative parameters $p$ an $n$. We find that a change in $p$ or $n$ results in a change in $z_0$ or $z_1$ respectively, while being invariant to changes in other variables. This is manifested in a MIG of 0.62, denoting moderate to strong disentanglement. $z_2$ and $z_3$ do not show correlations with either $p$ or $n$, emphasizing their "non-utilization". This shows that the latent variables of our model correctly discover the dimensionality 2 of the underlying generative procedure of ER graphs.

We repeat the experiment on a uni-, bi- and tri-parametric random graph model and two real-world graphs presented in the appendix. The selected graphs are complete binary tree graphs, \textit{BA graphs} \cite{barabasi99emergence}, \textit{Small-World graphs} \cite{watts1998collective} as well as the \textit{CORA} \cite{mccallum2017cora} and \textit{Wikipedia Hyperlink} \cite{rossi2015the} graph.

\subsection{Measuring Graph Topology-Node Attribute Dependence}
\label{sect:eval-attributes}

In addition to pure graph topology $\tau$, we consider node-level attributes $\Omega$ and measure the degree to which $\tau$ and $\Omega$ are mutually dependent. For example in a co-authorship graph where nodes represent authors and undirected links represent joint papers between authors, each node may hold additional information about the author's overall citation count. We denote this additional information as node attributes $\Omega$. Intuitively, more collaborations and therefore a higher node degree encourage a higher citation count, though there may be numerous other hidden correlations between graph topology and node attributes. Most existing topology-based approaches cannot make a general statement to what extent graph topology and node attributes are correlated without hand-picking particular topological properties such as the node degree \cite{zaki2000scalable}. 

We claim that the dependence between topological structure $\tau$ and attributes $\Omega$ is encoded in the latent variables. If $\tau$ and $\Omega$ are generated by independent generative procedures, they may be described by two disentangled sets of latent variables \cite{kim2018disentangling}. Proposing a node attribute randomization approach, we work with two data sets, the original graphs $X$ and their attribute-randomized versions $X^{\Delta \Omega}$. Since random graph generators such as ER graphs \cite{erdos1959on} do not cover node attributes, we first have to generate synthetic node attributes. Independent from $n$ and $p$, and hence from the topology $\tau$, all nodes of an ER graph are uniformly at random assigned the same node attribute which is a value between 0 and 1. We train the modified $\beta$-VAE on this graph data set $X$. After training, we randomize the node attributes, ending up with the randomized graph data $X^{\Delta \Omega}$. We vary the randomization degree $\Delta \Omega$ between 0 and 1, which denotes the fraction of randomized nodes. Finally, we present $X$ and $X^{\Delta \Omega}$ to the trained model in order to observe how the randomization affects the latent variables $z$.

\begin{figure}[t]
\centering
\includegraphics[width=1.0\textwidth]{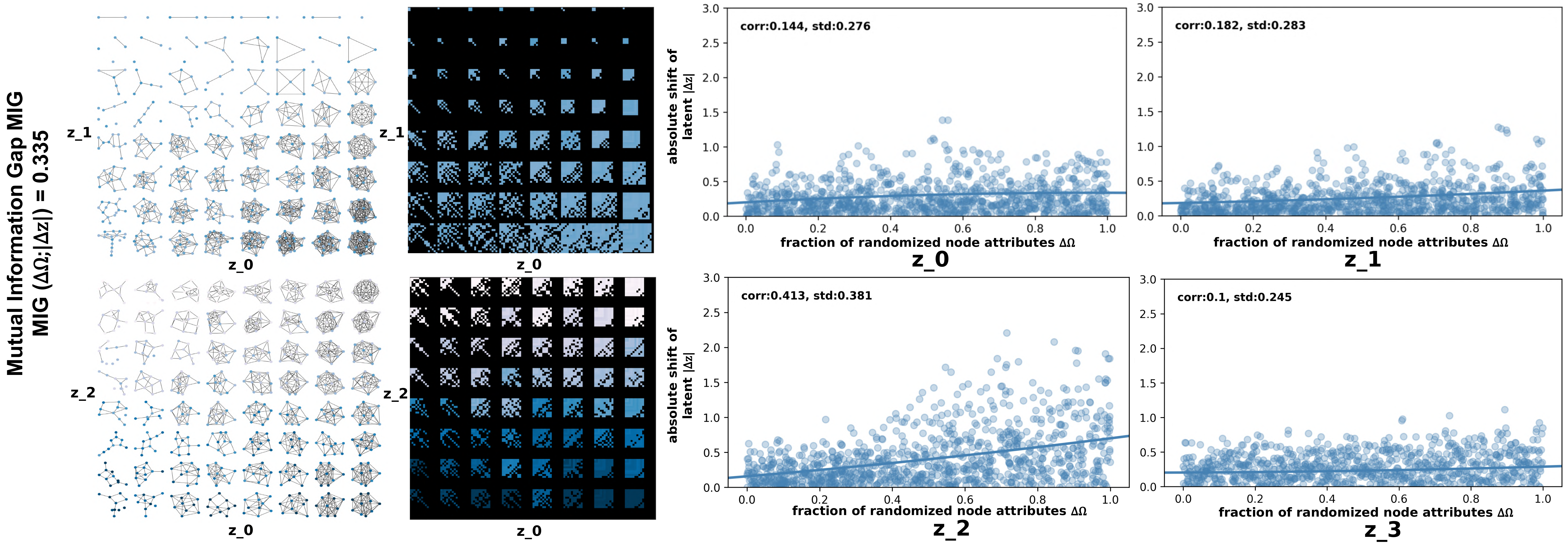}
\caption[Latent representation of ER graphs with uniform node attributes]{\textbf{Latent representation of ER graphs with uniform node attributes} \quad Node attribute values are indicated by the shade of blue. Traversing $z_0$ and $z_1$ while keeping other latent variables fix reveals a change in the topology $\tau$, as $p$ and $n$ vary. $z_0$ and $z_1$ are invariant to node attributes~$\Omega$. Since $z_2$ is most volatile to $\Delta \Omega$, it presumably models $\Omega$.}
\label{fig:figure_2}
\end{figure}

In the case of $\tau$-$\Omega$ independence, randomizing node attributes causes a shift in only those latent variables modelling $\Omega$. To indirectly quantify the dependence between $\tau$ and $\Omega$, we measure the correlation between $\Delta \Omega$ and $|\Delta z|$. $|\Delta z|$ describes the absolute change of $z_j$ due to $\Delta \Omega$. If only one latent variable changes while others are invariant, $\tau$ and $\Omega$ are generated from a fixed number of independent factors of variation \cite{chen2018isolating}. Disentanglement between latent variables serves as a proxy for the dependence of generative parameters $v_k$. Figure \ref{fig:figure_2} (left and center) displays manifolds of samples from latent space. Traversing $z_0$ and $z_1$ while fixing other latent variables reveals a change in $\tau$, as $p$ and $n$ change, but invariance to $\Omega$. $z_2$ is modelling $\Omega$, which is supported by figure \ref{fig:figure_2} (right) showing absolute shifts $\Delta z_j$ in the latents depending on the fraction of randomized nodes $\Delta \Omega$.

Treating the randomization degree $\Delta \Omega$ as a generative parameter, we calculate the mutual information (MI) between $\Delta \Omega$ and the absolute change in every latent $z_j$. $MIG ( \Delta \Omega;\Delta z| )$ then computes the gap between the first and second highest MI, normalized by the entropy $H(\Delta \Omega)$. In the equation below, $j^{max} = argmax_{j} MI(\Delta \Omega;|\Delta z|)$ denotes the index of latent $z_{j^{max}}$ with highest MI regarding $\Delta \Omega$.
\begin{align*}
&MIG(\Delta\Omega; \Delta|z_j|) = \frac{1}{H(\Delta\Omega)} \Big(MI(\Delta\Omega; \Delta|z_{j^{max}}|)  - \max_{j \neq j^{max}} MI(\Delta\Omega; \Delta|z_j|)  \Big)
\end{align*}

The latent variable reacting most strongly to $\Delta \Omega$ is $z_2$. $MIG ( \Delta \Omega; |\Delta z| )$ corresponding to figure \ref{fig:figure_2} is 0.335, indicating moderate disentanglement of $\Omega$ and $\tau$ as $z_{/2}$ are mostly invariant to $\Delta \Omega$. We repeat the experiment on the \textit{Microsoft Academic Graph (MAG)} \cite{sinha2015an} and \textit{Amazon Co-Purchasing Graph} \cite{leskovec2007the}, presented in the appendix. In particular for the  Microsoft Academic Graph, the analysis reveals a strong impact of the collaboration patterns (graph topology) on the citation count (node attributes).

\paragraph{Conclusion} This work demonstrates the potential of latent variable disentanglement in graph deep learning for unsupervised discovery of generative parameters of random and real-world graphs. Experiments have largely confirmed our hypotheses, but also revealed shortcomings. Future work should advance node order-independent graph decoders and target interpretability by exploiting generative models that do not sacrifice reconstruction fidelity for disentanglement.

\bibliography{cite}

\begin{thebibliography}{40}
\providecommand{\natexlab}[1]{#1}
\providecommand{\url}[1]{\texttt{#1}}
\expandafter\ifx\csname urlstyle\endcsname\relax
  \providecommand{\doi}[1]{doi: #1}\else
  \providecommand{\doi}{doi: \begingroup \urlstyle{rm}\Url}\fi

\bibitem[Barabási and Pósfai(2016)]{barabasi2016network}
Albert-László Barabási and Márton Pósfai.
\newblock \emph{Network Science}.
\newblock Cambridge University Press, Cambridge, 2016.

\bibitem[Chakrabarti and Faloutsos(2006)]{chakrabarti2006graph}
Deepayan Chakrabarti and Christos Faloutsos.
\newblock Graph mining: Laws, generators, and algorithms.
\newblock \emph{ACM Computing Survey}, 38, 2006.

\bibitem[Erd\"{o}s and R\'{e}nyi(1959)]{erdos1959on}
Paul Erd\"{o}s and Alfred R\'{e}nyi.
\newblock On random graphs.
\newblock \emph{Publicationes Mathematicae Debrecen}, 6, 1959.

\bibitem[Leskovec et~al.(2010)Leskovec, Chakrabarti, Kleinberg, Faloutsos, and
  Ghahramani]{leskovec2010kronecker}
Jure Leskovec, Deepayan Chakrabarti, Jon Kleinberg, Christos Faloutsos, and
  Zoubin Ghahramani.
\newblock Kronecker graphs: An approach to modeling networks.
\newblock \emph{Journal of Machine Learning Research}, 11, 2010.

\bibitem[Dorogovtsev and Mendes(2002)]{dorogovtsev2002evolution}
S.N. Dorogovtsev and J.F.F. Mendes.
\newblock Evolution of networks.
\newblock \emph{Advances in Physics}, 51\penalty0 (4), 2002.

\bibitem[Leskovec et~al.(2007{\natexlab{a}})Leskovec, Kleinberg, and
  Faloutsos]{leskovec2007graph}
Jure Leskovec, Jon Kleinberg, and Christos Faloutsos.
\newblock Graph evolution: Densification and shrinking diameters.
\newblock \emph{ACM Transactions on Knowledge Discovery from Data}, 8:\penalty0
  56--68, 2007{\natexlab{a}}.

\bibitem[Vazquez et~al.(2003)Vazquez, Flammini, Maritan, and
  Vespignani]{vazquez2003a}
Alexei Vazquez, Andrea Flammini, Andrea Maritan, and Alessandro Vespignani.
\newblock A global protein function prediction in protein-protein interaction
  networks.
\newblock \emph{Nature Biotech}, pages 697--700, 2003.

\bibitem[Zhou and Mondrag\'on(2004)]{zhou2004accurately}
Shi Zhou and Ra\'ul~J. Mondrag\'on.
\newblock Accurately modeling the internet topology.
\newblock \emph{Physical Review E}, 128:\penalty0 578--586, 2004.

\bibitem[Menczer(2004)]{menczer2004evolution}
Filippo Menczer.
\newblock Evolution of document networks.
\newblock \emph{Proceedings of the National Academy of Sciences}, 101:\penalty0
  5261--5265, 2004.

\bibitem[Bronstein et~al.(2017)Bronstein, {Bruna Estrach}, LeCun, Szlam, and
  Vandergheynst]{bronstein2017geometric}
{Michael M.} Bronstein, Joan {Bruna Estrach}, Yann LeCun, Arthur Szlam, and
  Pierre Vandergheynst.
\newblock Geometric deep learning: Going beyond euclidean data.
\newblock \emph{IEEE Signal Processing Magazine}, 34\penalty0 (4):\penalty0
  18--42, 2017.

\bibitem[Brugere et~al.(2018)Brugere, Gallagher, and
  Berger-Wolf]{brugere2018network}
Ivan Brugere, Brian Gallagher, and Tanya~Y. Berger-Wolf.
\newblock Network structure inference, a survey: Motivations, methods, and
  applications.
\newblock \emph{ACM Computing Survey}, 51\penalty0 (2), 2018.

\bibitem[Wang et~al.()Wang, Cui, and Zhu]{wang2016structural}
Daixin Wang, Peng Cui, and Wenwu Zhu.
\newblock Structural deep network embedding.
\newblock In \emph{Knowledge Discovery and Data Mining (KDD)}.

\bibitem[Johnson(2017)]{johnson2017learning}
Daniel~D. Johnson.
\newblock Learning graphical state transitions.
\newblock In \emph{International Conference on Learning Representations
  (ICLR)}, volume~34, pages 370--378, 2017.

\bibitem[You et~al.(2018)You, Ying, Ren, Hamilton, and
  Leskovec]{you2018graphrnn}
Jiaxuan You, Rex Ying, Xiang Ren, William~L. Hamilton, and Jure Leskovec.
\newblock Graphrnn: {A} deep generative model for graphs.
\newblock \emph{CoRR}, abs/1802.08773, 2018.

\bibitem[Li et~al.(2018)Li, Vinyals, Dyer, Pascanu, and
  Battaglia]{li2018learning}
Yujia Li, Oriol Vinyals, Chris Dyer, Razvan Pascanu, and Peter Battaglia.
\newblock Learning deep generative models of graphs.
\newblock In \emph{International Conference on Machine Learning (ICML)}, 2018.

\bibitem[Liu et~al.()Liu, Allamanis, Brockschmidt, and
  Gaunt]{liu2018constrained}
Qi~Liu, Miltiadis Allamanis, Marc Brockschmidt, and Alexander Gaunt.
\newblock Constrained graph variational autoencoders for molecule design.
\newblock In \emph{Advances in Neural Information Processing Systems
  (NeurIPS)}.

\bibitem[Li et~al.(2016)Li, Zemel, Brockschmidt, and Tarlow]{li2016gated}
Yujia Li, Richard Zemel, Marc Brockschmidt, and Daniel Tarlow.
\newblock Gated graph sequence neural networks.
\newblock In \emph{International Conference on Learning Representations
  (ICLR)}, 2016.

\bibitem[Simonovsky and Komodakis(2018)]{simonovsky2018graphvae}
Martin Simonovsky and Nikos Komodakis.
\newblock Graphvae: Towards generation of small graphs using variational
  autoencoders.
\newblock In \emph{Artificial Neural Networks and Machine Learning (ICANN)},
  pages 412--422. Springer International Publishing, 2018.

\bibitem[Ma et~al.()Ma, Cui, Kuang, Wang, and Zhu]{ma2019disentangled}
Jianxin Ma, Peng Cui, Kun Kuang, Xin Wang, and Wenwu Zhu.
\newblock Disentangled graph convolutional networks.
\newblock In \emph{International Conference on Machine Learning (ICML)}.

\bibitem[Bojchevski et~al.(2018)Bojchevski, Shchur, Z{\"u}gner, and
  G{\"u}nnemann]{bojchevski2018netgan}
Aleksandar Bojchevski, Oleksandr Shchur, Daniel Z{\"u}gner, and Stephan
  G{\"u}nnemann.
\newblock {N}et{GAN}: Generating graphs via random walks.
\newblock In \emph{International Conference on Machine Learning (ICML)}, 2018.

\bibitem[Grover and Leskovec(2016)]{grover2016node2vec}
Aditya Grover and Jure Leskovec.
\newblock Node2vec: Scalable feature learning for networks.
\newblock In \emph{Knowledge Discovery and Data Mining (KDD)}, 2016.

\bibitem[Cao et~al.()Cao, Lu, and Xu]{cao2016deep}
Shaosheng Cao, Wei Lu, and Qiongkai Xu.
\newblock Deep neural networks for learning graph representations.
\newblock In \emph{AAAI Conference on Artificial Intelligence (AAAI)}.

\bibitem[Perozzi et~al.(2016)Perozzi, Kulkarni, and
  Skiena]{walklets2016perozzi}
Bryan Perozzi, Vivek Kulkarni, and Steven Skiena.
\newblock Walklets: Multiscale graph embeddings for interpretable network
  classification.
\newblock \emph{CoRR}, abs/1605.02115, 2016.

\bibitem[Noutahi et~al.(2019)Noutahi, Beani, Horwood, and
  Tossou]{noutahi2019towards}
Emmanuel Noutahi, Dominique Beani, Julien Horwood, and Prudencio Tossou.
\newblock Towards interpretable sparse graph representation learning with
  laplacian pooling.
\newblock \emph{CoRR}, abs/1905.11577, 2019.

\bibitem[Navlakha(2017)]{navlakha2017learning}
Saket Navlakha.
\newblock Learning the structural vocabulary of a network.
\newblock \emph{Neural Computing}, 29\penalty0 (2):\penalty0 287--312, 2017.
\newblock ISSN 0899-7667.

\bibitem[{Bengio} et~al.(2013){Bengio}, {Courville}, and
  {Vincent}]{bengio2013representation}
Y.~{Bengio}, A.~{Courville}, and P.~{Vincent}.
\newblock Representation learning: A review and new perspectives.
\newblock \emph{IEEE Transactions on Pattern Analysis and Machine
  Intelligence}, 35\penalty0 (8):\penalty0 1798--1828, 2013.

\bibitem[Chen et~al.(2018)Chen, Li, Grosse, and Duvenaud]{chen2018isolating}
Tian~Qi Chen, Xuechen Li, Roger~B Grosse, and David~K Duvenaud.
\newblock Isolating sources of disentanglement in variational autoencoders.
\newblock In \emph{Advances in Neural Information Processing Systems
  (NeurIPS)}, pages 2610--2620, 2018.

\bibitem[Chen et~al.(2016)Chen, Duan, Houthooft, Schulman, Sutskever, and
  Abbeel]{chen2016infogan}
Xi~Chen, Yan Duan, Rein Houthooft, John Schulman, Ilya Sutskever, and Pieter
  Abbeel.
\newblock Infogan: Interpretable representation learning by information
  maximizing generative adversarial nets.
\newblock In D.~D. Lee, M.~Sugiyama, U.~V. Luxburg, I.~Guyon, and R.~Garnett,
  editors, \emph{Advances in Neural Information Processing Systems (NeurIPS)},
  pages 2172--2180. Curran Associates, Inc., 2016.

\bibitem[Kim and Mnih(2018)]{kim2018disentangling}
Hyunjik Kim and Andriy Mnih.
\newblock Disentangling by factorising.
\newblock \emph{Proceedings of the 35th International Conference on Machine
  Learning (ICML)}, 80:\penalty0 2649--2658, 2018.

\bibitem[Stühmer et~al.(2019)Stühmer, Turner, and
  Nowozin]{stuehmer2019independent}
Jan Stühmer, Richard~E. Turner, and Sebastian Nowozin.
\newblock Independent subspace analysis for unsupervised learning of
  disentangled representations.
\newblock \emph{arXiv}, abs/1909.05063, 2019.

\bibitem[Higgins et~al.(2017)Higgins, Matthey, Pal, Burgess, Glorot, Botvinick,
  Mohamed, and Lerchner]{higgins2017beta}
Irina Higgins, Lo{\"i}c Matthey, Arka Pal, Christopher Burgess, Xavier Glorot,
  Matthew Botvinick, Shakir Mohamed, and Alexander Lerchner.
\newblock beta-vae: Learning basic visual concepts with a constrained
  variational framework.
\newblock In \emph{International Conference on Learning Representations
  (ICLR)}, volume~1, pages 370--378, 2017.

\bibitem[Kipf and Welling(2017)]{kipf2017semi}
Thomas~N. Kipf and Max Welling.
\newblock Semi-supervised classification with graph convolutional networks.
\newblock \emph{International Conference on Learning Representations (ICLR)},
  34:\penalty0 34--42, 2017.

\bibitem[Guo et~al.(2006)Guo, Chen, and Zhou]{guo2006bosam}
Yuchun Guo, Changjia Chen, and Shi Zhou.
\newblock Bosam: A topology visualisation tool for large-scale complex
  networks.
\newblock \emph{arXiv preprint cs/0602034}, 2006.

\bibitem[Barabasi and Albert(1999)]{barabasi99emergence}
Albert-Laszlo Barabasi and Reka Albert.
\newblock Emergence of scaling in random networks.
\newblock \emph{Science}, 286, 1999.

\bibitem[Watts and Strogatz(1998)]{watts1998collective}
Duncan~J. Watts and Steven~H. Strogatz.
\newblock {Collective dynamics of 'small-world' networks}.
\newblock \emph{Nature}, 393\penalty0 (6684):\penalty0 440--442, 1998.

\bibitem[McCallum(2017)]{mccallum2017cora}
Andrew McCallum.
\newblock Cora dataset, 2017.

\bibitem[Rossi and Ahmed(2015)]{rossi2015the}
Ryan~A. Rossi and Nesreen~K. Ahmed.
\newblock The network data repository with interactive graph analytics and
  visualization.
\newblock In \emph{AAAI}, 2015.
\newblock URL \url{http://networkrepository.com}.

\bibitem[{Zaki}(2000)]{zaki2000scalable}
M.~J. {Zaki}.
\newblock Scalable algorithms for association mining.
\newblock \emph{IEEE Transactions on Knowledge and Data Engineering},
  12\penalty0 (3):\penalty0 372--390, 2000.

\bibitem[Sinha et~al.(2015)Sinha, Shen, Song, Ma, Eide, Hsu, and
  Wang]{sinha2015an}
Arnab Sinha, Zhihong Shen, Yang Song, Hao Ma, Darrin Eide, Bo-June~(Paul) Hsu,
  and Kuansan Wang.
\newblock An overview of microsoft academic service (mas) and applications.
\newblock In \emph{Proceedings of the 24th International Conference on World
  Wide Web (WWW)}, pages 243--246, New York, NY, USA, 2015. ACM.

\bibitem[Leskovec et~al.(2007{\natexlab{b}})Leskovec, Adamic, and
  Huberman]{leskovec2007the}
Jure Leskovec, Lada~A. Adamic, and Bernardo~A. Huberman.
\newblock The dynamics of viral marketing.
\newblock \emph{ACM Transactions on the Web}, 5:\penalty0 340--350,
  2007{\natexlab{b}}.

\end{thebibliography}

\newpage

\section*{Code}
To allow experimenting with the code, we provide an \href{https://colab.research.google.com/drive/1M--YX4dOSt3imDPdecPbjVX-T6Ae0_OG}{interactive notebook} at \url{https://colab.research.google.com/drive/1M--YX4dOSt3imDPdecPbjVX-T6Ae0_OG}.

\section*{Appendix}

\begin{figure}[h]
\centering
\includegraphics[width=1.0\textwidth]{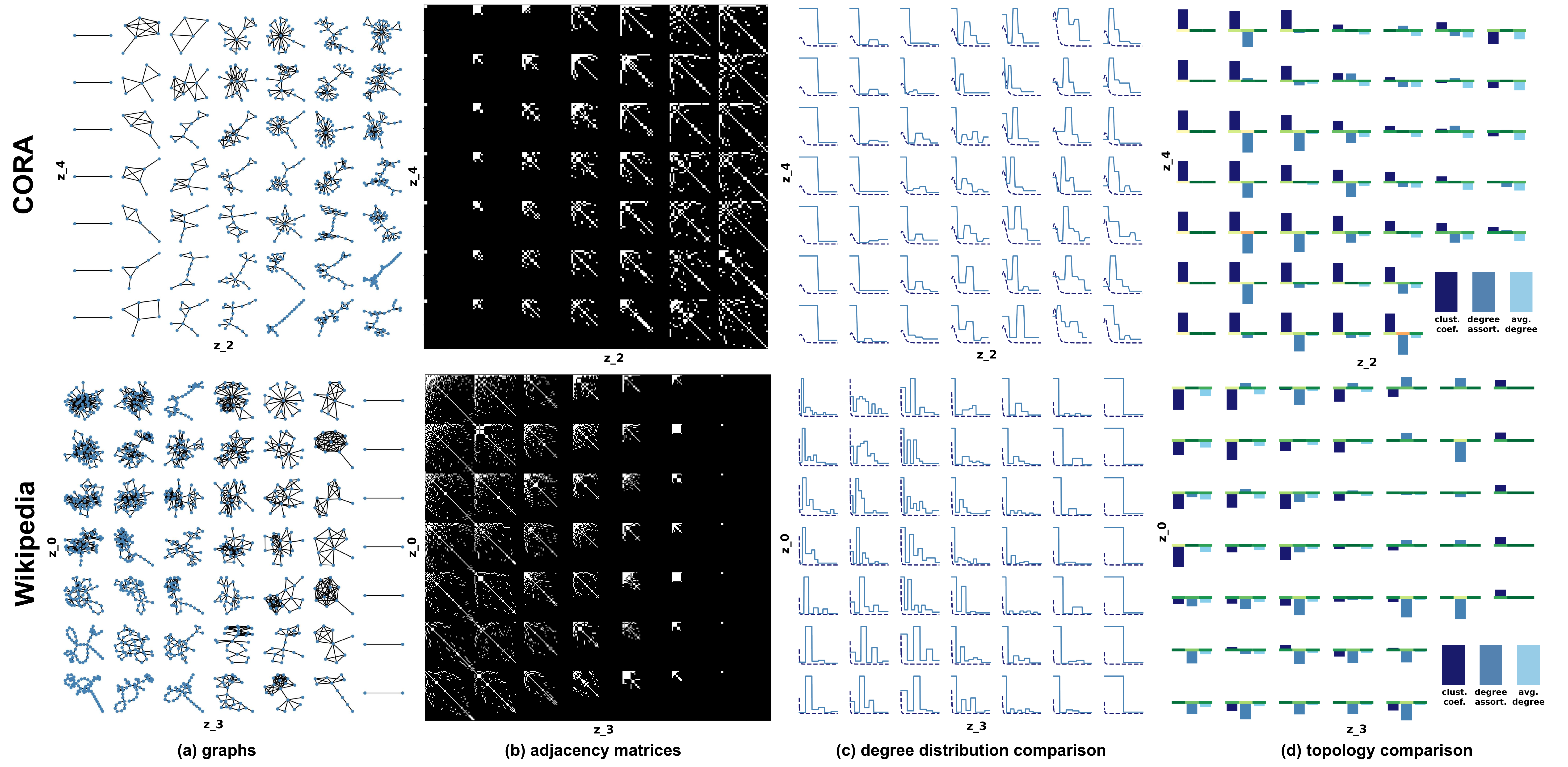}
\caption[Latent representations of real-world graphs]{\textbf{Latent representations of real-world graphs} \quad Latent space of \textit{$\beta$-VAE} model trained on 10,000 sub-graphs from \textit{CORA} \cite{mccallum2017cora} and \textit{Wikipedia} \cite{rossi2015the} sampled using \textit{Biased Second-Order Random Walks} \cite{grover2016node2vec}. Plot (a) and (b) show manifolds of decoded $x$ instances, presented as graphs and adjacency matrices respectively. Plot (c) compares the normalized degree distribution of $x$ with the distribution of the entire, original graph. Similarly, plot (d) shows the difference in clustering coefficient, degree assortativity and average degree.}
\label{fig:appendix_2}
\end{figure}

\begin{figure}[h]
\centering
\includegraphics[width=1.0\textwidth]{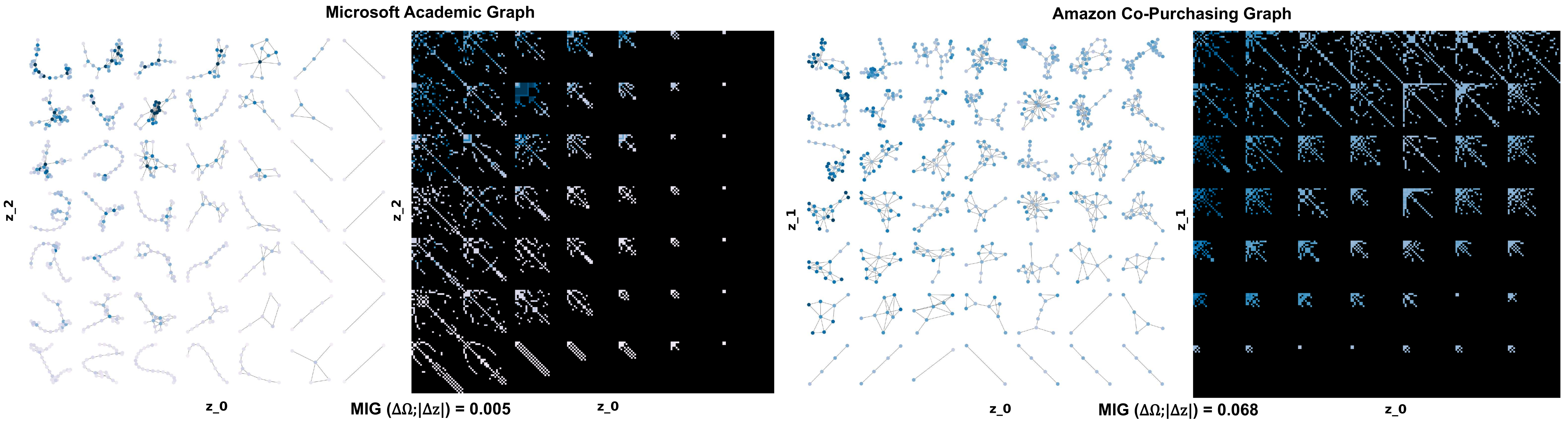}
\caption[Latent representations of real-world graphs with node attributes]{\textbf{Latent representations of real-world graphs with node attributes} \quad Manifold of graph instances obtained from traversing latent variables $z_j$ and decoding samples according to $p_{\Theta}(x|z)$. In the \textit{Microsoft Academic Graph}, topology $\tau$ and node attributes $\Omega$ can hardly be disentangled ($MIG \big(\Delta \Omega; |\Delta z|\big) = 0.005$). A reason lies in a strong correlation (0.4662) between the number of collaborations ($\tau$) and citations ($\Omega$).}
\label{fig:appendix_3}
\end{figure}

\begin{figure}[h]
\centering
\includegraphics[width=0.92\textwidth]{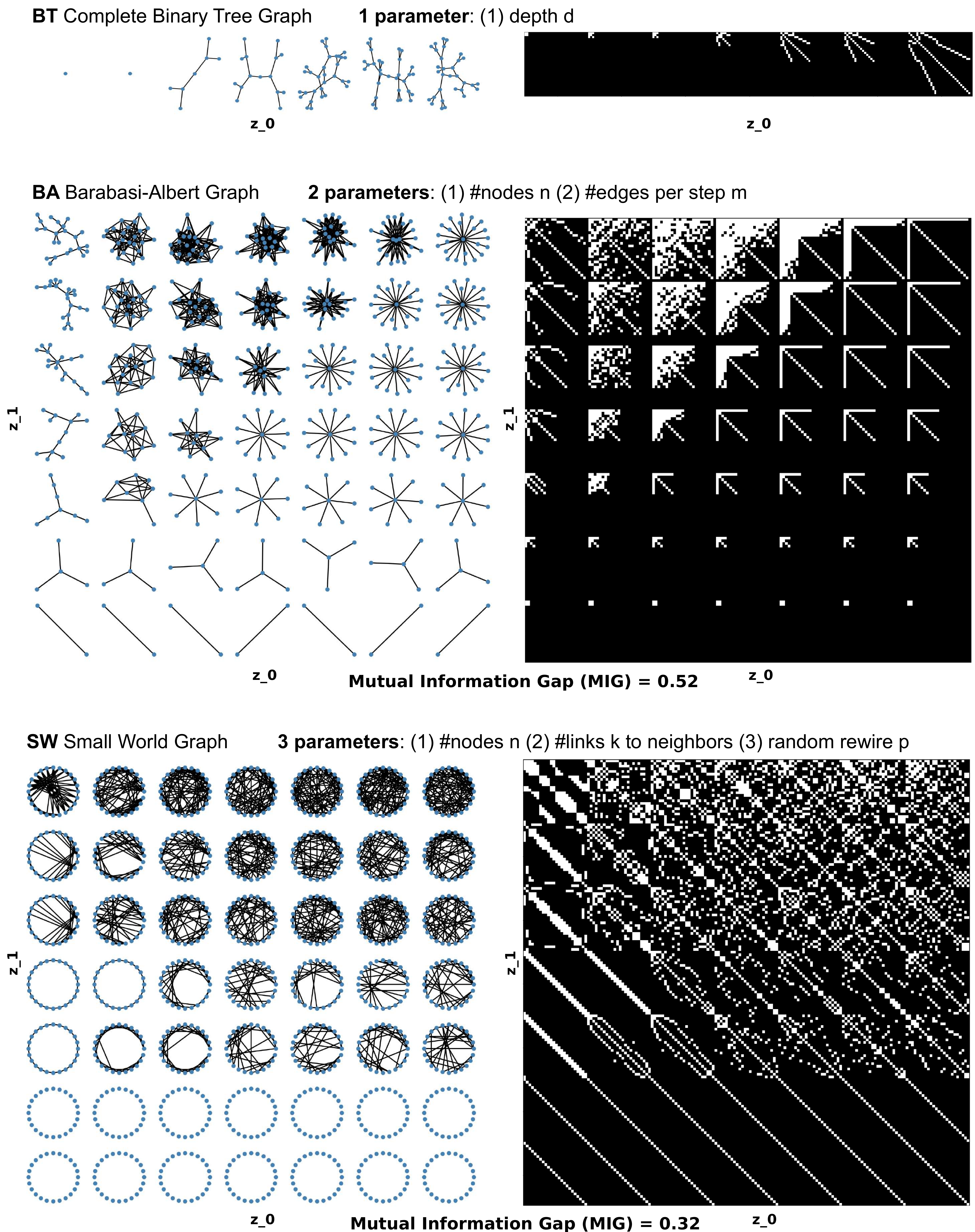}
\caption[Disentangled latent representation of uni-, bi- and tri-parametric random graph generator models]{\textbf{Disentangled latent representation of uni-, bi- and tri-parametric random graph generator models} \quad Latent representation of uni-parametric complete binary tree graph, bi-parametric \textit{Barabasi-Albert (BA) graphs} \cite{barabasi99emergence} and tri-parametric \textit{Small-World graphs (SW)} \cite{watts1998collective}. For visualizing the tri-parametric \textit{SW graphs}, we pick a fixed value for $z_0$ throughout all samples from the latent space. Since $z_0$ models the number of nodes $n$, all generated graphs in the manifolds are of fixed size. In compliance with intuition, the higher the degree of freedom in terms of generative parameters, the more difficult their successful disentanglement, manifested in a lower \textit{MIG} value for the tri-parametric \textit{SW graph}. If a uni-parametric model is described by a single latent variable, \textit{MIG} is not informative.}
\label{fig:appendix_1}
\end{figure}

\end{document}